\documentclass[lettersize,journal]{IEEEtran}
\usepackage{amsmath,amsfonts}
\usepackage{algorithmic}
\usepackage{array}
\usepackage[caption=false,font=normalsize,labelfont=sf,textfont=sf]{subfig}
\usepackage{textcomp}
\usepackage{stfloats}
\usepackage{url}
\usepackage{verbatim}
\usepackage{graphicx}
\usepackage{multirow}
\usepackage{booktabs}
\usepackage{makecell}
\usepackage{xcolor}
\def\BibTeX{{\rm B\kern-.05em{\sc i\kern-.025em b}\kern-.08em
    T\kern-.1667em\lower.7ex\hbox{E}\kern-.125emX}}
\usepackage{balance}
\begin{document}
\title{Low-Level Matters: An Efficient Hybrid Architecture for Robust Multi-frame Infrared Small Target Detection}

\author{Zhihua Shen, Siyang Chen, Han Wang, Tongsu Zhang, Xiaohu Zhang, Xiangpeng Xu and Xia Yang
\thanks{Zhihua Shen, Siyang Chen, Han Wang, Tongsu Zhang, Xiaohu Zhang, Xiangpeng Xu and Xia Yang are with the School of Aeronautics and Astronautics, Sun Yat-sen University, Guangzhou 510275, China (e-mail: yangxia7@mail.sysu.edu.cn).}
\thanks{Manuscript received xxxx xx, xxxx; revised xxxxx xx, xxxx. \textit{(Zhihua Shen, Siyang Chen contributed equally to this work. Corresponding author: Xiaohu Zhang, Xiangpeng Xu and Xia Yang.)}}}

\markboth{Journal of \LaTeX\ Class Files,~Vol.~18, No.~9, September~2020}%
{How to Use the IEEEtran \LaTeX \ Templates}

\maketitle

\begin{abstract}
	Multi-frame infrared small target detection (IRSTD) plays a crucial role in low-altitude and maritime surveillance. The hybrid architecture combining CNNs and Transformers shows great promise for enhancing multi-frame IRSTD performance. In this paper, we propose LVNet, a simple yet powerful hybrid architecture that redefines low-level feature learning in hybrid frameworks for multi-frame IRSTD. Our key insight is that the standard linear patch embeddings in Vision Transformers are insufficient for capturing the scale-sensitive local features critical to infrared small targets. To address this limitation, we introduce a multi-scale CNN frontend that explicitly models local features by leveraging the local spatial bias of convolution. Additionally, we design a U-shaped video Transformer for multi-frame spatiotemporal context modeling, effectively capturing the motion characteristics of targets. Experiments on the publicly available datasets IRDST and NUDT-MIRSDT demonstrate that LVNet outperforms existing state-of-the-art methods. Notably, compared to the current best-performing method, LMAFormer, LVNet achieves an improvement of 5.63\% / 18.36\% in nIoU, while using only 1/221 of the parameters and 1/92 / 1/21 of the computational cost. Ablation studies further validate the importance of low-level representation learning in hybrid architectures. Our code and trained models are available at https://github.com/ZhihuaShen/LVNet.
\end{abstract}


\section{Introduction}
\IEEEPARstart{I}{nfrared} detection offers all-weather capability, strong anti-interference performance, and long-range sensing, making it reliable in complex environments. It is widely used in military reconnaissance, maritime surveillance, and precision guidance\cite{zhao2022single,strickland2023infrared,yi2023spatial}. However, infrared small target detection (IRSTD) remains challenging due to the typically small size\cite{chen2022multi}, irregular shape\cite{lin2023learning,lin2024learning}, and low signal-to-clutter ratio (SCR)\cite{liu2023infrared} of infrared targets in complex backgrounds.

Single-frame detection methods primarily rely on the local saliency of the target relative to the background for recognition\cite{wang2019miss,zhang2024learning}. However, in complex scenes, such methods are susceptible to background interference, leading to missed detections and false alarms. In infrared imaging applications, moving targets such as unmanned aerial vehicles (UAVs), ships, and aircraft are the primary objects of detection\cite{aibibu2023efficient}, and their motion information in videos provides additional discriminative cues. Therefore, fully leveraging the motion characteristics of targets is crucial for improving detection performance. Compared to single-frame IRSTD, multi-frame detection methods can exploit temporal information to enhance the discriminability between the target and background, effectively suppress background interference, and improve the stability and accuracy of detection. These multi-frame IRSTD methods show broader application prospects\cite{kumar2023small}, particularly in critical tasks such as UAV tracking and maritime surveillance.

\begin{figure}
	\centering
	\includegraphics[width=\linewidth]{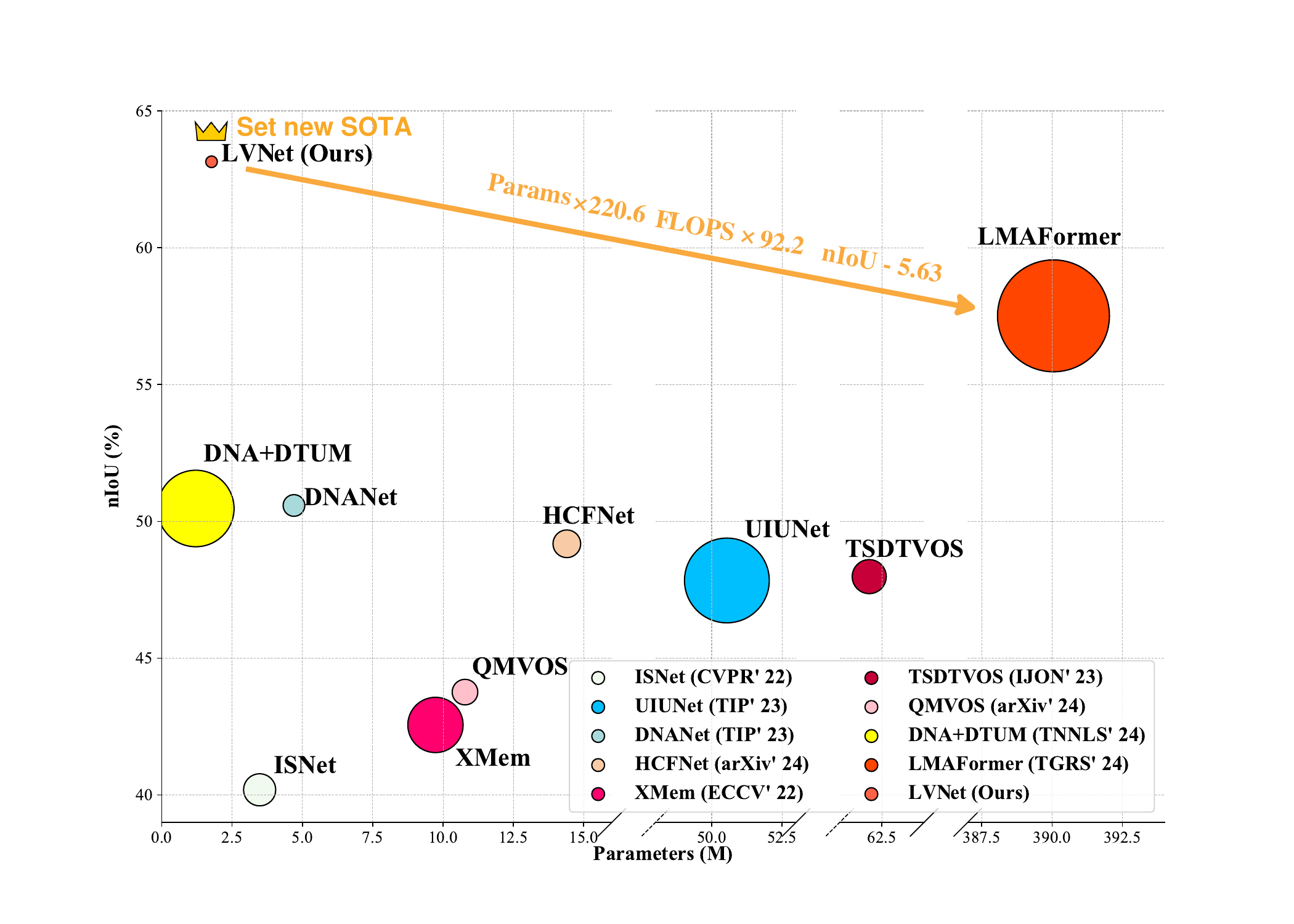}
	\caption{Comparison of the proposed LVNet with SOTA methods on the IRDST dataset\cite{sun2023receptive}. The area of the circles represents FLOPs. Our LVNet achieves a remarkable balance between computational efficiency and detection performance, setting a new SOTA.
		\label{fig:fig1}}
\end{figure}

Multi-frame IRSTD algorithms can be broadly categorized into two types: model-driven traditional methods and data-driven deep learning methods. Traditional methods mainly include background modeling-based\cite{zhang2023dim,zhang2024novel} and low-rank-based approaches\cite{liu2020small,zhang2020edge,zhang2023infrared,li2023sparse}. These methods are typically built on certain prior assumptions. However, in real-world scenarios, the constantly changing background and target may not fully satisfy these assumptions. Consequently, traditional methods often suffer from limitations in robustness and stability during practical applications.

Compared to traditional methods, deep learning-based approaches exhibit greater robustness and generalization capabilities, making them widely adopted in multi-frame IRSTD. Broadly, these methods can be classified into three main categories. The first category comprises CNN-based methods\cite{du2021spatial,yan2023stdmanet,li2023direction}, which capture the temporal motion characteristics of targets by either constructing a spatiotemporal feature tensor via channel-wise stacking or employing 3D convolutions. However, CNN-based methods are constrained by the limited receptive field of convolutional kernels, making it challenging to capture large-scale motion features. The second category consists of hybrid CNN-RNN methods\cite{chen2024sstnet,chen2024towards,chen2024convolutional}, which leverage RNNs to perform temporal modeling on the spatial features extracted by CNNs. However, for segmentation tasks, RNNs may struggle to effectively capture complex spatial relationships. Additionally, both CNN-based and CNN-RNN methods typically require image registration before processing multi-frame data. However, traditional registration techniques often introduce significant errors when applied to infrared images with limited texture information. The third category includes hybrid CNN-Transformer models\cite{vaswani2017attention,huang2024lmaformer}, which exhibit strong spatiotemporal modeling capabilities.

Although CNN-Transformer hybrid architectures have achieved promising detection performance in multi-frame IRSTD\cite{huang2024lmaformer}, existing methods predominantly leverage CNNs within the Transformer backbone to process high-level semantics. However, we argue that for IRSTD tasks, extracting low-level semantics using CNNs is even more critical.  

Infrared small targets typically exhibit low SCR, weak texture features, and are easily obscured by complex backgrounds. In such scenarios, while high-level semantics provide valuable global context, they are often insufficient for precise target localization. In contrast, low-level semantics capture rich edge, gradient, and local saliency, which enhance target visibility and reduce both missed and false detections. Furthermore, since infrared small targets are usually of limited scale and rely heavily on local features for differentiation, neglecting low-level semantics may result in the loss of crucial target information.  

Vision Transformer (ViT) employs a linear projection to generate patch embeddings, which lacks effective modeling of small-scale local features. To address this, we replace the linear projection layer in ViT's front end with CNNs, enabling more effective extraction of local target features. This design improves the model's ability to learn low-level representations without significantly increasing computational complexity.

Concretely, our contributions can be summarized as follows:
\begin{enumerate}
	\item[1)] For the first time, we introduce the concept of "low-level matters" within the CNN-Transformer hybrid architecture for multi-frame IRSTD. By integrating local bias into the patch embedding process, we significantly enhance the detection performance of the hybrid framework.
	\item[2)] We propose a U-shaped pure video Transformer backbone for spatiotemporal context modeling in multi-frame IRSTD tasks, enabling the model to maintain robust performance in dynamic backgrounds.
	\item[3)] Experimental results on the publicly available datasets NUDT-MIRSDT\cite{li2023direction} and IRDST\cite{sun2023receptive} demonstrate that our method outperforms SOTA approaches, achieving superior segmentation accuracy with a simple yet efficient architecture.
	\item[4)] The ablation study results demonstrate the importance of low-level representation learning for multi-frame IRSTD, providing insights for other moving small target detection tasks.
\end{enumerate}

\section{Related Work}
We review related work from two perspectives: multi-frame IRSTD and U-shaped Transformer architectures.

\subsection{Multi-frame IRSTD}
Historically, multi-frame IRSTD primarily relied on traditional methods based on prior knowledge, mainly including background modeling-based and low-rank-based approaches. Background modeling-based methods estimate the background using multiple frames. Zhao et al. \cite{zhao2023adaptive} employed spatial and temporal filters to suppress background noise, followed by thresholding for segmentation. Dang et al. \cite{dang2023infrared} utilized a radial basis function neural network to estimate spatiotemporal coupling coefficients and reconstruction functions for moving background modeling. Low-rank-based methods exploit the low-rank nature of the background and the sparsity of targets for detection. Liu et al. \cite{liu2023combining} applied a denoising neural network to construct an implicit regularizer, integrating low-rank priors with deep denoising priors. Wang et al. \cite{wang2021infrared} employed the tensor upper bound nuclear norm and non-overlapping patch-based spatiotemporal tensor modeling to detect potential targets. However, these methods often require background registration, and the estimation errors introduced in the registration process limit their performance in dynamic backgrounds.

In recent years, deep learning-based methods have garnered significant attention. Du et al. \cite{du2021spatial} proposed an end-to-end spatiotemporal feature extraction and target detection framework based on an inter-frame energy accumulation enhancement mechanism. Yan et al. \cite{yan2023stdmanet} designed a temporal multi-scale feature extractor to capture spatiotemporal features across different temporal scales, followed by a spatial multi-scale feature refinement module to enhance semantic representations. Chen et al. \cite{chen2024sstnet} employed CNN-LSTM nodes to capture spatiotemporal motion features within image tensor slices. Huang et al. \cite{huang2024lmaformer} introduced a multi-scale fusion Transformer encoder and a multi-frame joint query decoder for target detection.  

Currently, hybrid architectures combining CNNs and Transformers have demonstrated great potential in multi-frame IRSTD tasks. These architectures leverage the strengths of CNNs in extracting local target features while utilizing Transformers for global spatiotemporal context modeling. However, existing methods predominantly leverage CNNs within the Transformer backbone to process high-level semantics. We argue that for IRSTD tasks, extracting low-level semantics using CNNs is even more critical for hybrid architectures.

\subsection{U-shaped Transformer Architectures}
U-Net \cite{ronneberger2015u}, with its symmetric encoder-decoder architecture, has achieved remarkable success in medical image segmentation by effectively integrating multi-scale features. Since ViT \cite{dosovitskiy2020image} introduced the Transformer architecture into image classification, Transformers have led to significant advancements across various computer vision tasks. TransUnet \cite{chen2021transunet} combines the strengths of both Transformer and U-Net by leveraging Transformer modules for global context modeling of CNN-extracted features while maintaining a CNN-based encoder and skip connections for precise segmentation. Inspired by the Swin Transformer, Swin-Unet \cite{cao2022swin} constructs a pure Transformer-based symmetric architecture, demonstrating superior performance in medical image segmentation tasks. To extend the advantages of U-shaped Transformer architectures to multi-frame IRSTD, we propose a U-shaped pure video Transformer backbone.

\section{Method}
\begin{figure*}
	\centering
	\includegraphics[width=1\linewidth]{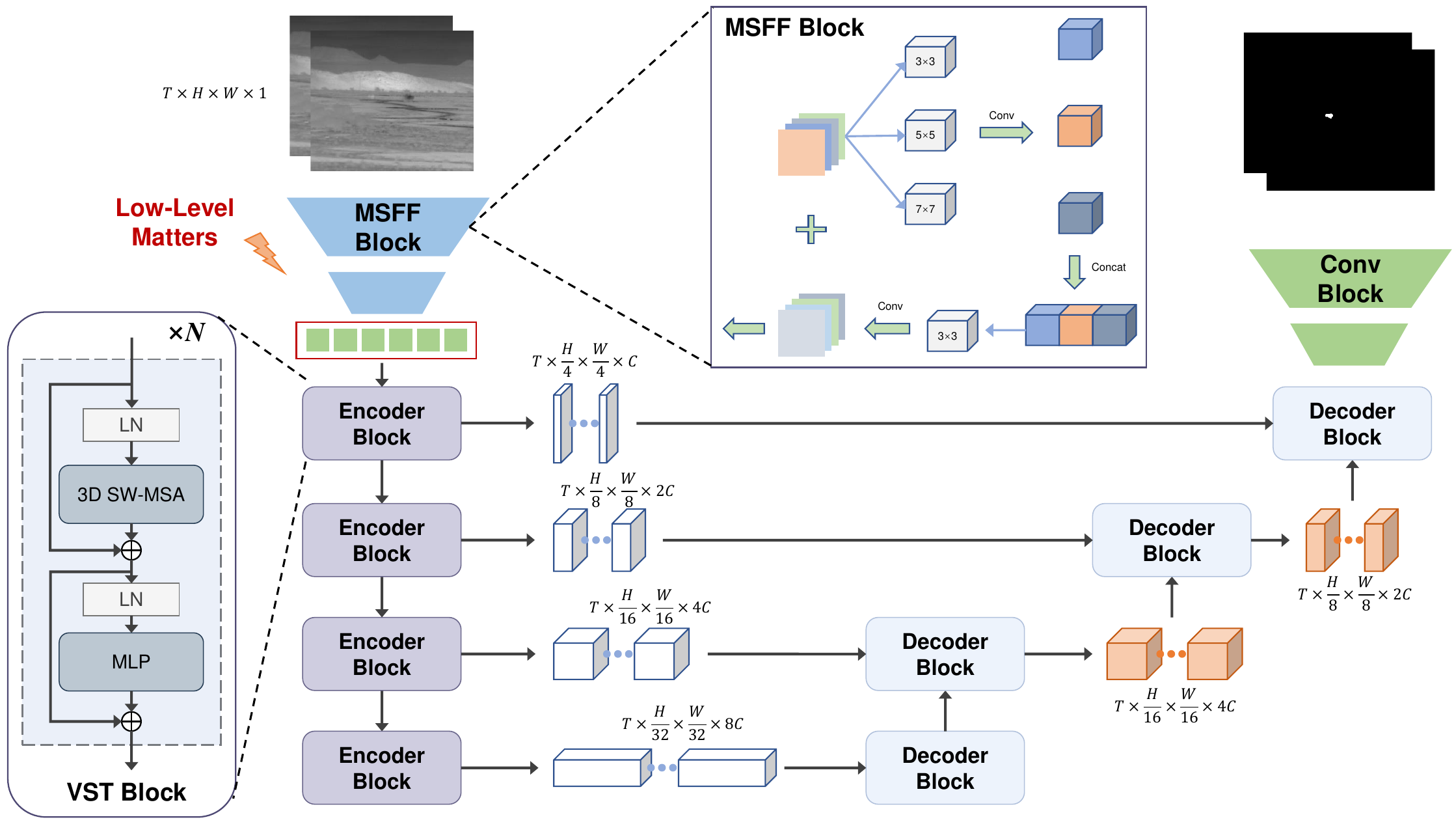}
	\caption{The architecture of LVNet.
		\label{fig:flowChat}}
\end{figure*}

The architecture of the proposed LVNet is shown in Fig. \ref{fig:flowChat}. LVNet consists of two key components: a shallow Convolutional U-Net (Conv U-Net) and a deep Video Transformer U-Net. The Conv U-Net is designed to enhance the local representation capability of patch embedding, while the Video Transformer U-Net focuses on spatiotemporal context modeling, effectively capturing temporal dependencies and global representations.

\subsection{Conv U-Net for Low-Level Feature Learning}

\textbf{Motivation} \ Low-level semantic features play a crucial role in IRSTD tasks, as they represent the target’s local characteristics and effectively enhance small target visibility in low-SCR and complex background scenarios. Since infrared small targets primarily rely on local features for differentiation, inadequate modeling of low-level semantics may lead to the loss of critical information, thereby affecting detection accuracy. Therefore, the model must efficiently extract and leverage low-level semantic features to improve target representation.

However, Vision Transformers (ViT) employ linear mapping in the generation of patch embeddings, which lacks effective modeling of local spatial features. To address this issue, we introduce CNNs into the patch embedding generation process. CNNs, with their local receptive fields and weight-sharing mechanisms, can effectively capture the fine-grained features of small targets while preserving their spatial structural information. Furthermore, to further improve the segmentation accuracy of small targets, we design the shallow network with a U-Net architecture that connects the input (patch embedding generation stage) and output (segmentation head) to enhance feature propagation and improve segmentation performance.

 The shape and size of infrared targets are variable, and their edges tend to be blurred. Moreover, weak targets are highly susceptible to interference from complex backgrounds. To overcome this challenge, we incorporate a multi-scale feature fusion (MSFF) module into the CNN structure, enabling the model to have receptive fields at various scales, thereby balancing the modeling of local details and the extraction of global background information.

\textbf{Details} \ To balance detection performance and computational efficiency, we designed a simple yet effective Conv U-Net. Its encoder consists of MSFF blocks and downsampling layers, while the decoder is composed of Conv Blocks and upsampling layers.

Specifically, the MSFF block employs parallel 3$\times$3, 5$\times$5, and 7$\times$7 convolution kernels to construct multi-scale receptive fields, thereby enhancing feature representation capability. The 3$\times$3 convolution primarily focuses on extracting edge features of the target, the 5$\times$5 convolution emphasizes the transition region between the target and the background, and the 7$\times$7 convolution helps model background context information for a more comprehensive feature representation. Afterward, a 3$\times$3 convolution is used to fuse the multi-scale features, further enhancing the target's representation ability. Residual connections are incorporated to maintain information flow, alleviate the gradient vanishing problem, and improve training stability. Notably, the Conv Block adopts a more lightweight structure, consisting only of a 3$\times$3 convolution layer with residual connections to reduce the parameter count and computational overhead.  

In the downsampling layer, we use a spatial-to-channel mapping approach to reduce the spatial resolution of the feature map while retaining key information as much as possible. This helps minimize spatial information loss and enhances the model's detection capability and robustness. The upsampling layer, conversely, performs the inverse operation to restore the spatial resolution, ensuring the complete reconstruction of target details.
 
 \subsection{Video Transformer U-Net for Spatiotemporal Context Modeling}
We select Video Swin Transformer (VST) as the encoder for spatiotemporal context modeling based on two key considerations. First, VST has demonstrated remarkable performance in temporal modeling tasks, with its sliding window mechanism enabling efficient processing of long-term sequential data and exhibiting strong temporal modeling capabilities. Second, its hierarchical structure allows it to accommodate targets moving at different speeds, thereby accurately capturing spatiotemporal features across varying motion states.  

However, unlike action recognition and temporal modeling tasks, IRSTD requires pixel-level target segmentation rather than merely recognizing overall temporal variations in the target. The U-shaped architecture, which fuses low-level high-resolution features from the encoder with high-level semantic information from the decoder, has been widely validated as an effective approach to mitigating spatial information loss caused by downsampling. Nevertheless, 2D conventional decoders primarily focus on spatial detail recovery while neglecting temporal consistency during the decoding stage, which compromises the integrity of spatiotemporal information.  

To address this, we design a fully symmetric decoder that mirrors the VST, ensuring that both spatial details are restored and temporal consistency is maintained, thereby enhancing the accuracy of IRSTD.

\section{Experiments}
\subsection{Setup}
\subsubsection{Datasets}

For the multi-frame IRSTD task, we utilized the widely used IRDST real dataset \cite{sun2023receptive} and the NUDT-MIRSDT \cite{li2023direction} simulated dataset. The IRDST dataset features a moving background, while the NUDT-MIRSDT dataset features a static background. The IRDST dataset comprises 41 training sequences and 17 validation sequences, with each sequence containing 50 images \cite{huang2024lmaformer}. The NUDT-MIRSDT dataset consists of 84 training sequences and 36 validation sequences, with each sequence including 100 images.
\subsubsection{Evaluation Metrics}
For the IRSDT task, we employed four widely used evaluation metrics, including intersection over union (IoU), normalized intersection over union (nIoU), probability of detection ($P_d$), and false alarm rate ($F_a$). IoU and nIoU are pixel-level metrics, defined as follows:
\begin{equation}
	IoU = \frac{TP}{T+P-TP},
\end{equation}
\begin{equation}
	nIoU = \frac{1}{N}\sum^N_i\frac{TP(i)}{T(i)+P(i)-TP(i)},
\end{equation}
where $N$ is the total number of samples, $T$, $P$ and $TP$ denote the pixel count of the ground truth, the prediction, and the true positive, respectively.
$P_d$ and $F_a$ are target-level metrics, defined as follows: 
\begin{equation}
	P_d = \frac{N_{true}}{N_{gt}},
\end{equation}
\begin{equation}
	F_a = \frac{N_{false}}{N_{all}},
\end{equation}
where $N_{true}$ represents the number of correctly detected targets, $N_{gt}$ denotes the total number of ground truth targets, $N_{false}$ is the number of falsely detected non-target pixels, and $N_{all}$ refers to the total pixels in the image. A predicted target is considered correctly detected if the distance between its center and the ground truth center is less than 3 pixels.

\subsubsection{Implementation Details}
The training process employed the Adam optimizer with an initial learning rate set to 1e-4. Specifically, the learning rate reduce by a factor of 0.9 when the loss stop improving. The minimum learning rate threshold was set to 1e-8. The networks are implemented in PyTorch and run on an Nvidia A100 GPU. The training batch size was set to 1, with a maximum of 200 epochs. No data augmentation strategies were applied, and the input images were single-channel images normalized by dividing by 255.

\subsection{Comparison to SOTA}
To evaluate the performance of our LVNet, we compared it with different methods. 

\subsubsection{Quantitative Evaluation}
\begin{figure*}
	\centering
	\includegraphics[width=\linewidth]{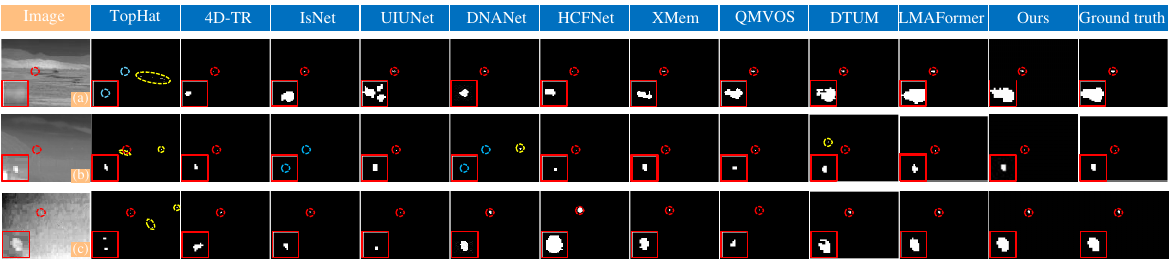}
	\caption{Visualization comparison of detection results via different methods on representative images from NUDT-MIRSDT dataset. Genuine targets are highlighted andmagnified in the lower left corner. Red circles are accurately detected targets, blue circles indicate missed detections, and yellow circles represent false alarms.
		\label{fig:Visualization comparison}}
\end{figure*}
\begin{table*}[]
	\centering
	
	\renewcommand\arraystretch{1.2} 
	\caption{Comparison to SOTA on IRDST and NUDT-MIRSDT Datasets. The magnitudes are Giga $(10^9)$ and Mega $(10^6)$ for FLOPs and Param respectively. The data is cited from \cite{huang2024lmaformer}.\label{table:quantitative evaluation}}
	\begin{tabular}{llcccccccccc}
		\Xhline{2\arrayrulewidth}
		\multicolumn{1}{c|}{\multirow{3}{*}{Method}} & \multicolumn{1}{c|}{\multirow{3}{*}{Publish}} & \multicolumn{1}{c|}{\multirow{3}{*}{Param}} & \multicolumn{1}{c|}{\multirow{3}{*}{FLOPs}} & \multicolumn{4}{c|}{IRDST}                                             & \multicolumn{4}{c}{NUDT-MIRSDT}                                         \\
		\multicolumn{1}{c|}{}                        & \multicolumn{1}{c|}{}                         & \multicolumn{1}{c|}{}                       & \multicolumn{1}{c|}{}                       & \multicolumn{2}{c}{Pixel-level} & \multicolumn{2}{c|}{Target-level}    & \multicolumn{2}{c}{Pixel-level} & \multicolumn{2}{c}{Target-level} \\
		\multicolumn{1}{c|}{}                        & \multicolumn{1}{c|}{}                         & \multicolumn{1}{c|}{}                       & \multicolumn{1}{c|}{}                       & IoU$\uparrow$            & nIoU$\uparrow$           & $P_d\uparrow$    & \multicolumn{1}{c|}{$F_a\downarrow$}      & IoU$\uparrow$            & nIoU$\uparrow$           & $P_d\uparrow$             & $F_a\downarrow$              \\
		\Xhline{2\arrayrulewidth}
		\multicolumn{12}{l}{\textit{Model-driven   \&\& Single Frame}}                                                                                                                                                                                                                                                                                  \\ \hline
		\multicolumn{1}{l|}{TopHat\cite{bai2010analysis}}                  & \multicolumn{1}{c|}{PR' 10}                   & \multicolumn{1}{c|}{\textendash}                      & \multicolumn{1}{c|}{\textendash}                      & 5.16           & 5.44           & 25.54 & \multicolumn{1}{c|}{378.82}  & 9.29            & 11.25         & 46.60          & 400.91          \\
		\multicolumn{1}{l|}{WSLCM\cite{han2020infrared}}                   & \multicolumn{1}{c|}{GRSL' 21}                 & \multicolumn{1}{c|}{\textendash}                      & \multicolumn{1}{c|}{\textendash}                      & 10.57          & 11.61          & 42.09 & \multicolumn{1}{c|}{\textcolor{blue}{\textbf{15.71}}}   & 7.22            & 7.83          & 51.79          & 29.64           \\
		\multicolumn{1}{l|}{RIPT\cite{dai2017reweighted}}                    & \multicolumn{1}{c|}{JSTARS' 17}               & \multicolumn{1}{c|}{\textendash}                      & \multicolumn{1}{c|}{\textendash}                      & 7.92           & 8.50           & 36.21 & \multicolumn{1}{c|}{85.49}   & 2.12            & 3.40          & 31.00          & 34.08           \\
		\multicolumn{1}{l|}{ANLPT\cite{zhang2023anlpt}}                   & \multicolumn{1}{c|}{RS' 23}                   & \multicolumn{1}{c|}{\textendash}                      & \multicolumn{1}{c|}{\textendash}                      & 3.01           & 3.76           & 24.94 & \multicolumn{1}{c|}{55.36}   & 2.05            & 2.35          & 14.97          & 15.50           \\ \hline
		\multicolumn{12}{l}{\textit{Model-driven   \&\& Multiple Frame}}                                                                                                                                                                                                                                                                                \\ \hline
		\multicolumn{1}{l|}{RCTVW\cite{liu2023representative}}                   & \multicolumn{1}{c|}{TGRS' 23}                 & \multicolumn{1}{c|}{\textendash}                      & \multicolumn{1}{c|}{\textendash}                      & 19.34          & 22.72          & 49.52 & \multicolumn{1}{c|}{1386.61} & 12.21           & 12.26         & 54.24          & 736.29          \\
		\multicolumn{1}{l|}{NFTDGSTV\cite{liu2023infrared}}                & \multicolumn{1}{c|}{TGRS' 23}                 & \multicolumn{1}{c|}{\textendash}                      & \multicolumn{1}{c|}{\textendash}                      & 5.37           & 10.50          & 31.37 & \multicolumn{1}{c|}{791.61}  & 11.55           & 11.94         & 69.81          & 123.28          \\
		\multicolumn{1}{l|}{STRL-LBCM\cite{luo2023spatial}}               & \multicolumn{1}{c|}{TAES' 23}                 & \multicolumn{1}{c|}{\textendash}                      & \multicolumn{1}{c|}{\textendash}                      & 20.85          & 25.70          & 58.27 & \multicolumn{1}{c|}{3165.20} & 5.13            & 5.25          & 46.16          & 707.90          \\
		\multicolumn{1}{l|}{SRSTT\cite{li2023sparse}}                   & \multicolumn{1}{c|}{TGRS' 23}                 & \multicolumn{1}{c|}{\textendash}                      & \multicolumn{1}{c|}{\textendash}                      & 19.07          & 23.18          & 46.28 & \multicolumn{1}{c|}{328.64}  & 16.19           & 15.90         & 65.28          & 141.75          \\
		\multicolumn{1}{l|}{4D-TR\cite{wu2023infrared}}                   & \multicolumn{1}{c|}{TGRS' 23}                 & \multicolumn{1}{c|}{\textendash}                      & \multicolumn{1}{c|}{\textendash}                      & 18.60          & 20.45          & 42.57 & \multicolumn{1}{c|}{94.40}   & 24.39           & 24.74         & 76.44          & 107.72          \\ \hline
		\multicolumn{12}{l}{\textit{Data-driven   \&\& Single Frame}}                                                                                                                                                                                                                                                                                   \\ \hline
		\multicolumn{1}{l|}{ISNet\cite{zhang2022isnet}}                   & \multicolumn{1}{l|}{CVPR' 22}                 & \multicolumn{1}{c|}{3.48}                   & \multicolumn{1}{c|}{31.38}                  & 38.62          & 40.19          & 84.29 & \multicolumn{1}{c|}{186.33}  & 61.33           & 62.40         & 68.74          & 40.26           \\
		\multicolumn{1}{l|}{UIUNet\cite{wu2022uiu}}                  & \multicolumn{1}{l|}{TIP' 23}                  & \multicolumn{1}{c|}{50.54}                  & \multicolumn{1}{c|}{218.00}                 & 47.50          & 47.84          & 86.93 & \multicolumn{1}{c|}{247.52}  & 59.37           & 61.54         & 64.16          & 19.04           \\
		\multicolumn{1}{l|}{DNANet\cite{li2022dense}}                  & \multicolumn{1}{l|}{TIP' 23}                  & \multicolumn{1}{c|}{4.70}                   & \multicolumn{1}{c|}{\textcolor{blue}{\textbf{14.28}}}                  & 49.11          & 50.58          & 91.61 & \multicolumn{1}{c|}{116.44}  & 49.48           & 58.96         & 71.02          & 179.04          \\
		\multicolumn{1}{l|}{HCFNet\cite{xu2024hcf}}                  & \multicolumn{1}{l|}{arXiv' 24}                & \multicolumn{1}{c|}{14.40}                  & \multicolumn{1}{c|}{23.28}                  & 48.56          & 49.18          & 90.05 & \multicolumn{1}{c|}{97.63}   & 58.91           & 64.48         & 73.36          & 26.73           \\ \hline
		\multicolumn{12}{l}{\textit{Data-driven   \&\& Multiple Frame}}                                                                                                                                                                                                                                                                                 \\ \hline
		\multicolumn{1}{l|}{XMem\cite{cheng2022xmem}}                    & \multicolumn{1}{l|}{ECCV' 22}                 & \multicolumn{1}{c|}{9.73}                   & \multicolumn{1}{c|}{92.81}                  & 42.66          & 42.56          & 78.54 & \multicolumn{1}{c|}{121.36}  & 70.28           & 70.54         & 93.97          & 8.96            \\
		\multicolumn{1}{l|}{TSDTVOS\cite{zhou2023tsdtvos}}                 & \multicolumn{1}{l|}{IJON' 23}                 & \multicolumn{1}{c|}{62.05}                  & \multicolumn{1}{c|}{35.38}                  & 48.09          & 47.98          & 87.05 & \multicolumn{1}{c|}{164.03}  & 56.89          & 57.50          & 73.94          & 217.95          \\
		\multicolumn{1}{l|}{QMVOS\cite{zhou2024video}}                   & \multicolumn{1}{l|}{arXiv' 24}                & \multicolumn{1}{c|}{10.78}                  & \multicolumn{1}{c|}{19.91}                  & 44.28          & 43.76          & 85.13 & \multicolumn{1}{c|}{160.60}  & 68.49           & 68.72         & 93.25          & 4.30            \\
		\multicolumn{1}{l|}{DNA+DTUM\cite{li2023direction}}                & \multicolumn{1}{l|}{TNNLS' 24}                & \multicolumn{1}{c|}{\textcolor{red}{\textbf{1.21}}}                   & \multicolumn{1}{c|}{177.74}                 & 49.69          & 50.47          & 88.01 & \multicolumn{1}{c|}{152.26}  & 70.76           & 72.35         & 95.33          & \textcolor{blue}{\textbf{2.54}}            \\
		\multicolumn{1}{l|}{LMAFormer\cite{huang2024lmaformer}}               & \multicolumn{1}{l|}{TGRS' 24}                 & \multicolumn{1}{c|}{390.05}                 & \multicolumn{1}{c|}{380.13}                 & \textcolor{blue}{\textbf{59.17}}          & \textcolor{blue}{\textbf{57.51}}          & \textcolor{red}{\textbf{99.64}} & \multicolumn{1}{c|}{\textcolor{red}{\textbf{14.95}}}   & \textcolor{blue}{\textbf{73.26}}           & \textcolor{blue}{\textbf{73.63}}         & \textcolor{red}{\textbf{99.68}}          & \textcolor{red}{\textbf{0.71}}            \\
		\multicolumn{1}{l|}{\textbf{LVNet (Ours)}}  & \multicolumn{1}{c|}{$\textendash$}                                              & \multicolumn{1}{c|}{\textcolor{blue}{\textbf{1.77}}}                   & \multicolumn{1}{c|}{\textcolor{red}{\textbf{4.12}}/\textcolor{blue}{\textbf{17.88}}}                  & \textcolor{red}{\textbf{65.93}}          & \textcolor{red}{\textbf{63.14}}          & \textcolor{blue}{\textbf{98.32}} & \multicolumn{1}{c|}{19.22}   & \textcolor{red}{\textbf{91.66}}           & \textcolor{red}{\textbf{91.99}}         & \textcolor{blue}{\textbf{98.82}}          & 6.28           \\ \Xhline{2\arrayrulewidth}
	\end{tabular}
\end{table*}

\begin{figure}
	\centering
	\includegraphics[width=\linewidth]{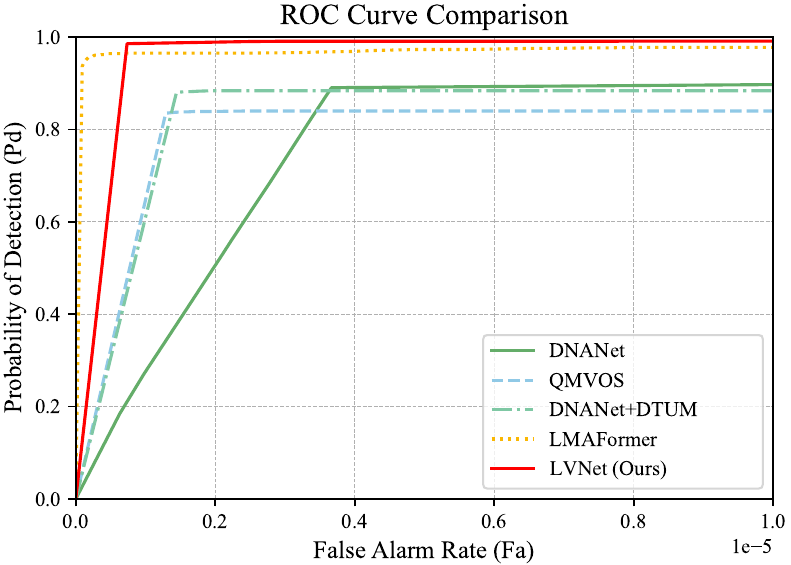}
	\caption{ROC curves of different methods on the NUDT-MIRSDT.
		\label{fig:ROC}}
\end{figure}
The quantitative evaluation results of different methods on two public datasets are shown in Table \ref{table:quantitative evaluation}. Overall, data-driven methods significantly outperform model-driven methods, and multi-frame approaches show a clear improvement over single-frame methods in terms of detection performance. This is mainly because data-driven methods can better extract infrared small target features and suppress background interference, while multi-frame methods can leverage temporal information to improve detection robustness and signal-to-noise ratio. Our proposed LVNet achieves outstanding results on both benchmark datasets, achieving the best performance in terms of IoU and nIoU, and comparable performance to SOTA methods in terms of $P_d$ and $F_a$.

In the IRDST dataset, due to strong background motion, some CNN-based multi-frame methods (e.g., DTUM) struggle to effectively leverage temporal information of both the background and the target because of the locality of convolution operations. This limitation often introduces more interference and errors, resulting in inferior detection performance compared to single-frame methods (e.g., DNANet and HCFNet). In contrast, LMAFormer and our LVNet leverage the powerful global modeling capability of transformers, making them more adaptable to background motion. This demonstrates that hybrid transformer-CNN architectures surpass pure CNN architectures in spatiotemporal modeling capability.  

As a transformer-CNN hybrid architecture, our method performs slightly worse than the SOTA LMAFormer in terms of $P_d$ and $F_a$ but exhibits significant advantages in pixel-level metrics. Across both datasets, our approach outperforms LMAFormer in IoU and nIoU by 6.76\%/18.40\% and 5.63\%/18.36\%, respectively. This indicates that our method achieves higher precision in pixel-level predictions for infrared small targets. This can be attributed to two key factors. First, we place greater emphasis on learning low-level semantic features, such as local saliency, within the hybrid architecture. By leveraging CNNs with multi-scale receptive fields, we enhance the local spatial features of the target. Second, we design a symmetric video transformer architecture to facilitate comprehensive multi-scale spatiotemporal feature interaction.
\subsubsection{Qualitative Evaluation}
The qualitative results shown in Fig. \ref{fig:Visualization comparison} intuitively compare the detection performance of different methods on representative images from the NUDT-MIRSDT dataset. As observed in the figure, model-driven methods and single-frame methods produce more missed detections (highlighted by blue circles) and false alarms (highlighted by yellow circles) compared to data-driven multi-frame methods, especially in scenarios with low SCR and complex backgrounds (e.g., subfigure (b)). Benefiting from the powerful feature representation learning capability of neural networks and the utilization of temporal information, data-driven multi-frame methods exhibit significant advantages in detection performance and segmentation accuracy. Among these methods, DTUM tends to have under-segmentation problems, while LMAFormer tends to have over-segmentation problems. Among all the algorithms presented, our method produces segmentation results closest to the ground truth, indicating that LVNet has surpassed existing SOTA methods (e.g., LMAFormer) in segmentation accuracy.

\subsubsection{Comparison of ROC Curves}
We plot the ROC curves of different algorithms on the NUDT-MIRSDT dataset, as shown in Fig. \ref{fig:ROC}. Under very low $F_a$ conditions, our method's $P_d$ is slightly lower than that of LMAFormer, but it achieves the highest $P_d$ under most $F_a$ conditions. This indicates that our method strikes a better balance between $P_d$ and $F_a$.
\subsubsection{Comparison of Computational Cost}
\begin{table*}[]
	\centering
	\renewcommand\arraystretch{1.2} 
	\caption{Ablation study on the temporal dimension of 3D tokens on IRDST and NUDT-MIRSDT.\label{table:ablation_combined}}
	\begin{tabular}{cc|cc|cccc|cccc}
		\Xhline{2\arrayrulewidth}
		&                               &                             &                             & \multicolumn{4}{c|}{IRDST}                                       & \multicolumn{4}{c}{NUDT-MIRSDT}                                \\
		\multirow{-2}{*}{T} & \multirow{-2}{*}{Window size} & \multirow{-2}{*}{Param}     & \multirow{-2}{*}{FLOPs}     & IoU    & nIoU   & Pd     & Fa     & IoU    & nIoU   & Pd     & Fa     \\ \hline
		2                   & 2$\times$7$\times$7          & 1.77 & 4.12  & \textbf{65.93} & \textbf{63.14} & \textbf{98.32} & 19.22  & 69.25      & 60.96      & 88.29      & 124.40      \\
		4                   & 4$\times$7$\times$7          & 1.77 & 8.47  & 63.49  & 60.93  & 96.88  & \textbf{11.15} & 88.38  & 87.95  & 98.27  & 13.60 \\
		8                   & 8$\times$7$\times$7          & 1.77 & 17.88 & -      & -      & -      & -      & \textbf{91.66} & \textbf{91.99} & \textbf{98.82} & \textbf{6.28}  \\
		\Xhline{2\arrayrulewidth}
	\end{tabular}
\end{table*}
The parameter count and computational complexity of different models are shown in Table \ref{table:quantitative evaluation}. Our LVNet has only 1.77M parameters, ranking second. Our best-performing model on the IRDST dataset uses a temporal dimension of 2 frames, achieving 4.12G FLOPs, ranking first. On the NUDT-MIRSDT dataset, our best model uses a temporal dimension of 8 frames, reaching 17.88G FLOPs, ranking second. Thanks to its lightweight architecture and the efficient use of CNNs, our LVNet demonstrates significant advantages in resource consumption and computational efficiency, even surpassing most single-frame models.  

Overall, although our method slightly underperforms LMAFormer in terms of $P_d$ and $F_a$, LMAFormer has 220.6 times more parameters and 92.3/21.3 times the computational complexity of our model. Clearly, our approach achieves a better balance between performance and resource efficiency, making it a more practical and valuable solution for multi-frame IRSDT tasks.

\subsection{Ablation Study}

\subsubsection{Temporal Dimension of 3D Tokens}
We perform an ablation study on the temporal dimension of 3D tokens, where the temporal dimension is equal to the size of the temporal window. The results are presented in Table \ref{table:ablation_combined}. Generally, a larger temporal dimension incurs higher computational costs and slower inference speed. On the IRDST dataset, increasing the temporal dimension led to a decline in detection performance. However, on the NUDT-MIRSDT dataset, it improved detection performance. We attribute this to the impact of background motion, which weakens the long-term motion features of the target. As a result, the NUDT-MIRSDT with static background benefits more from an increased temporal dimension.
\begin{table}[]
	\centering
	\renewcommand\arraystretch{1.2} 
	\caption{Ablation study on the temporal window size on NUDT-MIRSDT.\label{table:window size}}
	\begin{tabular}{cc|cc|cccc}
		\Xhline{2\arrayrulewidth}
		\multirow{2}{*}{T} & \multirow{2}{*}{Window size} & \multirow{2}{*}{Param} & \multirow{2}{*}{FLOPs} & \multicolumn{4}{c}{NUDT-MIRSDT} \\
		&                              &                        &                        & IoU    & nIoU   & Pd    & Fa    \\ \hline
		8                  & 8$\times$7$\times$7                          & 1.77                   & 17.88                  & \textbf{91.66}  & \textbf{91.99}  & \textbf{98.82} & \textbf{6.28}  \\
		8                  & 4$\times$7$\times$7                          & 1.77                   & 16.94                  & 89.3   & 88.97  & 98.79 & 16.56 \\
		8                  & 2$\times$7$\times$7                          & 1.77                   & 16.47                  & 74.65  & 69.97  & 92.3  & 86.1  \\ \Xhline{2\arrayrulewidth}
	\end{tabular}
\end{table}
\subsubsection{Temporal Window Size}
Fixing the temporal dimensionof 3D tokens to 8, we perform an ablation study over temporal window sizes of 2/4/8. The results reported in Table \ref{table:window size} indicate that the difference in FLOPs across different temporal window sizes is relatively small. However, as the temporal window size increases, all performance metrics improve significantly. Therefore, we set the window size to 8 as the default setting for NUDT-MIRSDT.
\subsubsection{Embedding Dims}
\begin{table}[]
	\centering
	\renewcommand\arraystretch{1.2} 
	\caption{Ablation study on the embedding dims on IRDST.\label{table:ablation 2 on IRDST}}
	\begin{tabular}{c|cc|cccc}
		\Xhline{2\arrayrulewidth}
		\multirow{2}{*}{C} & \multirow{2}{*}{Param} & \multirow{2}{*}{FLOPs} & \multicolumn{4}{c}{IRDST}     \\
		&                        &                        & IoU   & nIoU  & Pd    & Fa    \\ \hline
		12                 & 0.45                   & 1.16                   & 62.16 & 59.53 & 96.52 & 13.19 \\
		24                 & 1.77                   & 4.12                   & \textbf{65.93} & \textbf{63.14} & \textbf{98.32} & 19.22 \\
		48                 & 7.03                   & 15.79                  & 63.76 & 61.31 & 97.12 & \textbf{9.67}  \\ \Xhline{2\arrayrulewidth}
	\end{tabular}
\end{table}
We perform an ablation study on the embedding dims $C$, with the results on the IRDST dataset presented in Table \ref{table:ablation 2 on IRDST}. As shown in the table, compared to a embedding dims of 12, increasing the dims to 24 significantly improved detection performance, with nIoU and $P_d$ increasing by 3.61\% and 1.80\%, respectively, while the parameter count and computational cost only increased slightly. Although the model with 48 dims reduced the $F_a$ metric by 9.55$\times$$10^{-6}$, it exhibited performance degradation across other metrics. Therefore, we set the embedding dims to 24 as the default configuration.
\subsubsection{Layer Numbers}
\begin{table}[]
	\centering
	\renewcommand\arraystretch{1.2} 
	\caption{Ablation study on the layer numbers on IRDST.\label{table:ablation 3 on IRDST}}
	\begin{tabular}{c|cc|cccc}
		\Xhline{2\arrayrulewidth}
		\multirow{2}{*}{Layer numbers} & \multirow{2}{*}{Param} & \multirow{2}{*}{FLOPs} & \multicolumn{4}{c}{IRDST}     \\
		&                        &                        & IoU   & nIoU  & Pd    & Fa    \\ \hline
		\{1,1,1,1\}                               & 1.47                   & 3.34                   & 65.54 & 62.47 & 97.24 & 13.71 \\
		\{2,2,2,1\}                               & 1.77                   & 4.12                   & \textbf{65.93} & \textbf{63.14} & \textbf{98.32} & 19.22 \\
		\{3,3,3,1\}                               & 2.06                   & 4.90                    & 61.73 & 59.84 & \textbf{98.32} & \textbf{13.59} \\ \Xhline{2\arrayrulewidth}
	\end{tabular}
\end{table}
The ablation study on the layer numbers of transformer encoder is reported in Table \ref{table:ablation 3 on IRDST}. The model with a layer configuration of \{2,2,2,1\} achieved the best performance in terms of IoU, nIoU, and $P_d$. The model with a layer configuration of \{3,3,3,1\} achieved the best results for $P_d$ and $F_a$, but led to a significant decrease of 4.20\% and 3.3\% in IoU and nIoU, respectively. After considering all factors, we selected \{2,2,2,1\} as the default layer configuration.
\subsubsection{Effect of Conv U-Net}
\begin{table}[]
	\centering
	\renewcommand\arraystretch{1.2} 
	\caption{Ablation study on the Conv U-Net on IRDST.\label{table:ablation 4 on IRDST}}
	\begin{tabular}{c|cc|cccc}
		\Xhline{2\arrayrulewidth}
		\multirow{2}{*}{Conv U-Net}        & \multirow{2}{*}{Param}   & \multirow{2}{*}{FLOPs}    & \multicolumn{4}{c}{IRDST}                                                                                     \\
		&                          &                           & IoU                       & \multicolumn{1}{c}{nIoU}  & \multicolumn{1}{c}{Pd}    & \multicolumn{1}{c}{Fa}    \\ \hline
		w. Conv U-Net                      & 1.77                     & 4.12                      & \textbf{65.93}                     & \textbf{63.14} & \textbf{98.32} & 19.22 \\
		w/o Conv U-Net                     & 1.70  & 2.03 & 62.25 & 58.77                     & 96.76                     & 20.22                     \\
		MSFF$\rightarrow$Res block & 1.72 & 2.76 & 62.45 & 59.59                     & 95.56                     & \textbf{8.39}                      \\ \Xhline{2\arrayrulewidth}
	\end{tabular}
\end{table}
To verify the effectiveness of Conv U-Net, we conducted an ablation study, with the results on the IRDST dataset presented in Table \ref{table:ablation 4 on IRDST}. Compared to directly generating patch embeddings via linear mapping, it resulted in improvements of 3.68\%, 4.37\%, and 1.56\% in IoU, nIoU, and $P_d$, respectively. Compared to the residual block (res block), our MSFF block improved IoU, nIoU, and $P_d$ by 3.48\%, 3.55\%, and 2.76\%, respectively. The results indicate that enhancing the multi-scale local feature representation of patch embedding is effective in improving the detection performance of the hybrid architecture for infrared small targets.

\subsubsection{Effect of Transformer Decoder}
\begin{table}[]
	\centering
	\renewcommand\arraystretch{1.2} 
	\caption{Ablation study on the VST blocks of decoder on NUDT-MIRSDT.\label{table:ablation 5 on IRDST}}
	\begin{tabular}{c|cc|cccc}
		\Xhline{2\arrayrulewidth}
		\multirow{2}{*}{Decoder   block} & \multirow{2}{*}{Param} & \multirow{2}{*}{FLOPs} & \multicolumn{4}{c}{NUDT-MIRSDT}     \\
		&                        &                        & IoU   & nIoU  & Pd    & Fa    \\ \hline
		Conv2d block                      & 1.58                   & 15.12                   & 88.65 & 87.96 & 98.79 & 14.52  \\
		Conv3d block                      & 2.13                   & 20.29                   & 88.60 & 88.62 & 98.70 & 6.88 \\
		VST block                        & 1.77                   & 17.88                   & \textbf{91.66} & \textbf{91.99} & \textbf{98.82} & \textbf{6.28} \\ \Xhline{2\arrayrulewidth}
	\end{tabular}
\end{table}
The decoder we designed is mainly composed of VST blocks and patch expanding layers. To validate the effectiveness of the VST block in the decoder, a experiment on VST is conducted by replacing VSTs with 2D Convolutions (conv2d) and 3D ones (conv3d) to validate the performance gains. The results are shown in Table \ref{table:ablation 5 on IRDST}. Compared to the conv2d block and conv3d block, the VST block resulted in improvements of 3.01\%/3.06\% and 4.03\%/3.37\% in IoU and nIoU, respectively. This suggests that performing spatiotemporal interactions in the decoder is also beneficial, especially in reducing false alarms. It also validates the effectiveness of our designed video transformer U-Net.

In our Transformer Decoder, we apply a patch expanding layer corresponding to the patch merging layer to perform upsampling and feature dimension expansion. To verify the effectiveness of the patch expanding layer, we ablated three commonly used upsampling designs: bilinear interpolation (Bilinear), transpose convolution (TransConv), and the patch expanding layer. The ablation study results for upsampling are reported in Table \ref{table:ablation 6 on IRDST}. Compared to Bilinear and TransConv, the patch expanding layer improved the pixel-level nIoU by 1.70\% and 4.67\%, respectively. The results indicate that the LVNet combined with the patch expanding layer can achieve better segmentation accuracy.
\begin{table}[]
	\centering
	\renewcommand\arraystretch{1.2} 
	\caption{Ablation study on the up-sampling of decoder on IRDST.\label{table:ablation 6 on IRDST}}
	\begin{tabular}{c|cc|cccc}
		\Xhline{2\arrayrulewidth}
		\multirow{2}{*}{Up-sampling} & \multirow{2}{*}{Param} & \multirow{2}{*}{FLOPs} & \multicolumn{4}{c}{IRDST}     \\
		&                        &                        & IoU   & nIoU  & Pd    & Fa    \\ \hline
		Bilinear      & 1.68                   & 4.07                   & 64.43 & 61.44 & 97.72 & 13.75 \\
		TransConv       & 1.74                   & 4.19                   & 61.26 & 58.47 & 96.52 & \textbf{8.67}  \\
		Patch expand                 & 1.77                   & 4.12                   & \textbf{65.93} & \textbf{63.14} & \textbf{98.32} & 19.22 \\ \Xhline{2\arrayrulewidth}
	\end{tabular}
\end{table}

\section{Conclusion}
In this paper, we propose a simple yet powerful U-shaped hybrid architecture for robust multi-frame IRSTD. We emphasize the importance of effective low-level representation learning in hybrid architectures. A multi-scale CNN is introduced for generating patch embeddings, though other CNN structures could be substituted. Our key focus is to highlight the necessity of incorporating CNNs at this stage. Additionally, we design a pure Transformer-based video U-Net to facilitate spatiotemporal interactions among the extracted multi-frame features. Experiments on public datasets demonstrate that the proposed method outperforms other SOTA approaches. With low resource consumption and high computational efficiency, our method holds great promise for multi-frame IRSTD. The emphasis on "low-level matters" in this paper may also provide valuable insights for other moving small target detection tasks.

\bibliographystyle{IEEEtran}
\bibliography{reference}

\begin{thebibliography}{10}
\providecommand{\url}[1]{#1}
\csname url@samestyle\endcsname
\providecommand{\newblock}{\relax}
\providecommand{\bibinfo}[2]{#2}
\providecommand{\BIBentrySTDinterwordspacing}{\spaceskip=0pt\relax}
\providecommand{\BIBentryALTinterwordstretchfactor}{4}
\providecommand{\BIBentryALTinterwordspacing}{\spaceskip=\fontdimen2\font plus
\BIBentryALTinterwordstretchfactor\fontdimen3\font minus \fontdimen4\font\relax}
\providecommand{\BIBforeignlanguage}[2]{{%
\expandafter\ifx\csname l@#1\endcsname\relax
\typeout{** WARNING: IEEEtran.bst: No hyphenation pattern has been}%
\typeout{** loaded for the language `#1'. Using the pattern for}%
\typeout{** the default language instead.}%
\else
\language=\csname l@#1\endcsname
\fi
#2}}
\providecommand{\BIBdecl}{\relax}
\BIBdecl

\bibitem{zhao2022single}
M.~Zhao, W.~Li, L.~Li, J.~Hu, P.~Ma, and R.~Tao, ``Single-frame infrared small-target detection: A survey,'' \emph{IEEE Geoscience and Remote Sensing Magazine}, vol.~10, no.~2, pp. 87--119, 2022.

\bibitem{strickland2023infrared}
R.~N. Strickland, ``Infrared techniques for military applications,'' in \emph{Infrared Methodology and Technology}.\hskip 1em plus 0.5em minus 0.4em\relax CRC Press, 2023, pp. 397--427.

\bibitem{yi2023spatial}
H.~Yi, C.~Yang, R.~Qie, J.~Liao, F.~Wu, T.~Pu, and Z.~Peng, ``Spatial-temporal tensor ring norm regularization for infrared small target detection,'' \emph{IEEE Geoscience and Remote Sensing Letters}, vol.~20, pp. 1--5, 2023.

\bibitem{chen2022multi}
Y.~Chen, L.~Li, X.~Liu, and X.~Su, ``A multi-task framework for infrared small target detection and segmentation,'' \emph{IEEE Transactions on Geoscience and Remote Sensing}, vol.~60, pp. 1--9, 2022.

\bibitem{lin2023learning}
F.~Lin, S.~Ge, K.~Bao, C.~Yan, and D.~Zeng, ``Learning shape-biased representations for infrared small target detection,'' \emph{IEEE Transactions on Multimedia}, 2023.

\bibitem{lin2024learning}
F.~Lin, K.~Bao, Y.~Li, D.~Zeng, and S.~Ge, ``Learning contrast-enhanced shape-biased representations for infrared small target detection,'' \emph{IEEE Transactions on Image Processing}, 2024.

\bibitem{liu2023infrared}
F.~Liu, C.~Gao, F.~Chen, D.~Meng, W.~Zuo, and X.~Gao, ``Infrared small and dim target detection with transformer under complex backgrounds,'' \emph{IEEE Transactions on Image Processing}, vol.~32, pp. 5921--5932, 2023.

\bibitem{wang2019miss}
H.~Wang, L.~Zhou, and L.~Wang, ``Miss detection vs. false alarm: Adversarial learning for small object segmentation in infrared images,'' in \emph{Proceedings of the IEEE/CVF International Conference on Computer Vision}, 2019, pp. 8509--8518.

\bibitem{zhang2024learning}
G.~Zhang, G.~Xu, H.~Wang, S.~Chen, Y.~Shan, and X.~Zhang, ``Learning dynamic local context representations for infrared small target detection,'' \emph{arXiv preprint arXiv:2412.17401}, 2024.

\bibitem{aibibu2023efficient}
T.~Aibibu, J.~Lan, Y.~Zeng, W.~Lu, and N.~Gu, ``An efficient rep-style gaussian--wasserstein network: Improved uav infrared small object detection for urban road surveillance and safety,'' \emph{Remote Sensing}, vol.~16, no.~1, p.~25, 2023.

\bibitem{kumar2023small}
N.~Kumar and P.~Singh, ``Small and dim target detection in ir imagery: A review,'' \emph{arXiv preprint arXiv:2311.16346}, 2023.

\bibitem{sun2023receptive}
H.~Sun, J.~Bai, F.~Yang, and X.~Bai, ``Receptive-field and direction induced attention network for infrared dim small target detection with a large-scale dataset irdst,'' \emph{IEEE Transactions on Geoscience and Remote Sensing}, vol.~61, pp. 1--13, 2023.

\bibitem{zhang2023dim}
Y.~Zhang, X.~Chen, P.~Rao, and L.~Jia, ``Dim moving multi-target enhancement with strong robustness for false enhancement,'' \emph{Remote Sensing}, vol.~15, no.~19, p. 4892, 2023.

\bibitem{zhang2024novel}
G.~Zhang, H.~Zhang, Z.~Shen, D.~Kong, C.~Ning, F.~Shang, and X.~Zhang, ``A novel detection method for warhead fragment targets in optical images under dynamic strong interference environments,'' \emph{Defence Technology}, 2024.

\bibitem{liu2020small}
H.-K. Liu, L.~Zhang, and H.~Huang, ``Small target detection in infrared videos based on spatio-temporal tensor model,'' \emph{IEEE Transactions on Geoscience and Remote Sensing}, vol.~58, no.~12, pp. 8689--8700, 2020.

\bibitem{zhang2020edge}
P.~Zhang, L.~Zhang, X.~Wang, F.~Shen, T.~Pu, and C.~Fei, ``Edge and corner awareness-based spatial--temporal tensor model for infrared small-target detection,'' \emph{IEEE Transactions on Geoscience and Remote Sensing}, vol.~59, no.~12, pp. 10\,708--10\,724, 2020.

\bibitem{zhang2023infrared}
Z.~Zhang, P.~Gao, S.~Ji, X.~Wang, and P.~Zhang, ``Infrared small target detection combining deep spatial-temporal prior with traditional priors,'' \emph{IEEE Transactions on Geoscience and Remote Sensing}, 2023.

\bibitem{li2023sparse}
J.~Li, P.~Zhang, L.~Zhang, and Z.~Zhang, ``Sparse regularization-based spatial--temporal twist tensor model for infrared small target detection,'' \emph{IEEE Transactions on Geoscience and Remote Sensing}, vol.~61, pp. 1--17, 2023.

\bibitem{du2021spatial}
J.~Du, H.~Lu, L.~Zhang, M.~Hu, S.~Chen, Y.~Deng, X.~Shen, and Y.~Zhang, ``A spatial-temporal feature-based detection framework for infrared dim small target,'' \emph{IEEE Transactions on Geoscience and Remote Sensing}, vol.~60, pp. 1--12, 2021.

\bibitem{yan2023stdmanet}
P.~Yan, R.~Hou, X.~Duan, C.~Yue, X.~Wang, and X.~Cao, ``Stdmanet: Spatio-temporal differential multiscale attention network for small moving infrared target detection,'' \emph{IEEE transactions on geoscience and remote sensing}, vol.~61, pp. 1--16, 2023.

\bibitem{li2023direction}
R.~Li, W.~An, C.~Xiao, B.~Li, Y.~Wang, M.~Li, and Y.~Guo, ``Direction-coded temporal u-shape module for multiframe infrared small target detection,'' \emph{IEEE Transactions on Neural Networks and Learning Systems}, 2023.

\bibitem{chen2024sstnet}
S.~Chen, L.~Ji, J.~Zhu, M.~Ye, and X.~Yao, ``Sstnet: Sliced spatio-temporal network with cross-slice convlstm for moving infrared dim-small target detection,'' \emph{IEEE Transactions on Geoscience and Remote Sensing}, 2024.

\bibitem{chen2024towards}
S.~Chen, L.~Ji, S.~Zhu, M.~Ye, H.~Ren, and Y.~Sang, ``Towards dense moving infrared small target detection: New datasets and baseline,'' \emph{IEEE Transactions on Geoscience and Remote Sensing}, 2024.

\bibitem{chen2024convolutional}
S.~Chen, H.~Wang, Z.~Shen, K.~Wang, and X.~Zhang, ``Convolutional long-short term memory network for space debris detection and tracking,'' \emph{Knowledge-Based Systems}, vol. 304, p. 112535, 2024.

\bibitem{vaswani2017attention}
A.~Vaswani, ``Attention is all you need,'' \emph{Advances in Neural Information Processing Systems}, 2017.

\bibitem{huang2024lmaformer}
Y.~Huang, X.~Zhi, J.~Hu, L.~Yu, Q.~Han, W.~Chen, and W.~Zhang, ``Lmaformer: Local motion aware transformer for small moving infrared target detection,'' \emph{IEEE Transactions on Geoscience and Remote Sensing}, 2024.

\bibitem{zhao2023adaptive}
Y.~Zhao, Y.~Li, C.~Zhu, S.~Wang, Z.~Lan, and Y.~Qiao, ``An adaptive spatial-temporal local feature difference method for infrared small-moving target detection,'' in \emph{2023 8th IEEE International Conference on Network Intelligence and Digital Content (IC-NIDC)}.\hskip 1em plus 0.5em minus 0.4em\relax IEEE, 2023, pp. 346--351.

\bibitem{dang2023infrared}
C.~Dang, Z.~Li, C.~Hao, and Q.~Xiao, ``Infrared small marine target detection based on spatiotemporal dynamics analysis,'' \emph{Remote Sensing}, vol.~15, no.~5, p. 1258, 2023.

\bibitem{liu2023combining}
T.~Liu, Q.~Yin, J.~Yang, Y.~Wang, and W.~An, ``Combining deep denoiser and low-rank priors for infrared small target detection,'' \emph{Pattern Recognition}, vol. 135, p. 109184, 2023.

\bibitem{wang2021infrared}
G.~Wang, B.~Tao, X.~Kong, and Z.~Peng, ``Infrared small target detection using nonoverlapping patch spatial--temporal tensor factorization with capped nuclear norm regularization,'' \emph{IEEE Transactions on Geoscience and Remote Sensing}, vol.~60, pp. 1--17, 2021.

\bibitem{ronneberger2015u}
O.~Ronneberger, P.~Fischer, and T.~Brox, ``U-net: Convolutional networks for biomedical image segmentation,'' in \emph{Medical image computing and computer-assisted intervention--MICCAI 2015: 18th international conference, Munich, Germany, October 5-9, 2015, proceedings, part III 18}.\hskip 1em plus 0.5em minus 0.4em\relax Springer, 2015, pp. 234--241.

\bibitem{dosovitskiy2020image}
A.~Dosovitskiy, ``An image is worth 16x16 words: Transformers for image recognition at scale,'' \emph{arXiv preprint arXiv:2010.11929}, 2020.

\bibitem{chen2021transunet}
J.~Chen, Y.~Lu, Q.~Yu, X.~Luo, E.~Adeli, Y.~Wang, L.~Lu, A.~L. Yuille, and Y.~Zhou, ``Transunet: Transformers make strong encoders for medical image segmentation,'' \emph{arXiv preprint arXiv:2102.04306}, 2021.

\bibitem{cao2022swin}
H.~Cao, Y.~Wang, J.~Chen, D.~Jiang, X.~Zhang, Q.~Tian, and M.~Wang, ``Swin-unet: Unet-like pure transformer for medical image segmentation,'' in \emph{European conference on computer vision}.\hskip 1em plus 0.5em minus 0.4em\relax Springer, 2022, pp. 205--218.

\bibitem{bai2010analysis}
X.~Bai and F.~Zhou, ``Analysis of new top-hat transformation and the application for infrared dim small target detection,'' \emph{Pattern Recognition}, vol.~43, no.~6, pp. 2145--2156, 2010.

\bibitem{han2020infrared}
J.~Han, S.~Moradi, I.~Faramarzi, H.~Zhang, Q.~Zhao, X.~Zhang, and N.~Li, ``Infrared small target detection based on the weighted strengthened local contrast measure,'' \emph{IEEE Geoscience and Remote Sensing Letters}, vol.~18, no.~9, pp. 1670--1674, 2020.

\bibitem{dai2017reweighted}
Y.~Dai and Y.~Wu, ``Reweighted infrared patch-tensor model with both nonlocal and local priors for single-frame small target detection,'' \emph{IEEE journal of selected topics in applied earth observations and remote sensing}, vol.~10, no.~8, pp. 3752--3767, 2017.

\bibitem{zhang2023anlpt}
Z.~Zhang, C.~Ding, Z.~Gao, and C.~Xie, ``Anlpt: Self-adaptive and non-local patch-tensor model for infrared small target detection,'' \emph{Remote Sensing}, vol.~15, no.~4, p. 1021, 2023.

\bibitem{liu2023representative}
T.~Liu, J.~Yang, B.~Li, Y.~Wang, and W.~An, ``Representative coefficient total variation for efficient infrared small target detection,'' \emph{IEEE Transactions on Geoscience and Remote Sensing}, 2023.

\bibitem{luo2023spatial}
Y.~Luo, X.~Li, Y.~Yan, and C.~Xia, ``Spatial-temporal tensor representation learning with priors for infrared small target detection,'' \emph{IEEE Transactions on Aerospace and Electronic Systems}, 2023.

\bibitem{wu2023infrared}
F.~Wu, H.~Yu, A.~Liu, J.~Luo, and Z.~Peng, ``Infrared small target detection using spatiotemporal 4-d tensor train and ring unfolding,'' \emph{IEEE Transactions on Geoscience and Remote Sensing}, vol.~61, pp. 1--22, 2023.

\bibitem{zhang2022isnet}
M.~Zhang, R.~Zhang, Y.~Yang, H.~Bai, J.~Zhang, and J.~Guo, ``Isnet: Shape matters for infrared small target detection,'' in \emph{Proceedings of the IEEE/CVF Conference on Computer Vision and Pattern Recognition}, 2022, pp. 877--886.

\bibitem{wu2022uiu}
X.~Wu, D.~Hong, and J.~Chanussot, ``Uiu-net: U-net in u-net for infrared small object detection,'' \emph{IEEE Transactions on Image Processing}, vol.~32, pp. 364--376, 2022.

\bibitem{li2022dense}
B.~Li, C.~Xiao, L.~Wang, Y.~Wang, Z.~Lin, M.~Li, W.~An, and Y.~Guo, ``Dense nested attention network for infrared small target detection,'' \emph{IEEE Transactions on Image Processing}, vol.~32, pp. 1745--1758, 2022.

\bibitem{xu2024hcf}
S.~Xu, S.~Zheng, W.~Xu, R.~Xu, C.~Wang, J.~Zhang, X.~Teng, A.~Li, and L.~Guo, ``Hcf-net: Hierarchical context fusion network for infrared small object detection,'' \emph{arXiv preprint arXiv:2403.10778}, 2024.

\bibitem{cheng2022xmem}
H.~K. Cheng and A.~G. Schwing, ``Xmem: Long-term video object segmentation with an atkinson-shiffrin memory model,'' in \emph{European Conference on Computer Vision}.\hskip 1em plus 0.5em minus 0.4em\relax Springer, 2022, pp. 640--658.

\bibitem{zhou2023tsdtvos}
W.~Zhou, Y.~Zhao, F.~Zhang, B.~Luo, L.~Yu, B.~Chen, C.~Yang, and W.~Gui, ``Tsdtvos: Target-guided spatiotemporal dual-stream transformers for video object segmentation,'' \emph{Neurocomputing}, vol. 555, p. 126582, 2023.

\bibitem{zhou2024video}
H.~Zhou, R.~Hu, and X.~Li, ``Video object segmentation with dynamic query modulation,'' \emph{arXiv preprint arXiv:2403.11529}, 2024.

\end{thebibliography}

\end{document}